\documentclass[conference]{IEEEtran}
\IEEEoverridecommandlockouts
\usepackage{cite}
\usepackage{amsmath,amssymb,amsfonts}
\usepackage{algorithmic}
\usepackage{graphicx}
\usepackage{textcomp}
\usepackage{xcolor}
\usepackage{comment}
\usepackage{multirow}
\usepackage{makecell}
\usepackage{tabularx}

\def\BibTeX{{\rm B\kern-.05em{\sc i\kern-.025em b}\kern-.08em
    T\kern-.1667em\lower.7ex\hbox{E}\kern-.125emX}}
\begin{document}


\title{QLAM: A Quantum Long-Attention Memory Approach to Long-Sequence Token Modeling}

\author{
    Hoang-Quan Nguyen$^{1, 2}$, 
    Sankalp Pandey$^{1, 2}$,
    Khoa Luu$^{1, 2}$ \\
    $^1$Department of Electrical Engineering and Computer Science, University of Arkansas, AR \\
    $^2$Quantum AI Lab, University of Arkansas, AR \\
    \small{\texttt{\{hn016, sankalpp, khoaluu\}@uark.edu}} \
}

\maketitle

\begin{abstract}

Modeling long-range dependencies in sequential data remains a central challenge in machine learning. Transformers address this challenge through attention mechanisms, but their quadratic complexity with respect to sequence length limits scalability to long contexts. State-space models (SSMs) provide an efficient alternative with linear-time computation by evolving a latent state through recurrent updates, but their memory is typically formed via additive or linear transitions, which can limit their ability to capture complex global interactions across tokens. In this work, we introduce one of the first studies to leverage the superposition property of quantum systems to enhance state-based sequence modeling. In particular, we propose Quantum Long-Attention Memory (QLAM), a hybrid quantum-classical memory mechanism that can be viewed as a quantum extension of state-space models. Instead of maintaining a classical latent state updated through additive dynamics, QLAM represents the hidden state as a quantum state whose amplitudes encode a superposition of historical information. The state evolves through parameterized quantum circuits conditioned on the input, enabling a non-classical, globally update mechanism. In this way, QLAM preserves the recurrent and linear-time structure of SSMs while fundamentally enriching the memory representation through quantum superposition. Unlike attention mechanisms that explicitly compute pairwise interactions, QLAM implicitly captures global dependencies through the evolution of the quantum state, and retrieves task-relevant information via query-dependent measurements. We evaluate QLAM on sequential variants of standard image classification benchmarks, including sMNIST, sFashion-MNIST, and sCIFAR-10, where images are flattened into token sequences. Across all tasks, QLAM consistently improves over recurrent baselines and transformer-based models. These results suggest that QLAM provides a principled extension of state-space formulations, where quantum superposition replaces classical additive memory, enabling richer global representations while retaining favorable computational scaling. This opens a new research direction for sequence modeling by integrating quantum computation into the core design of memory mechanisms.




\end{abstract}


\section{Introduction}

The ability to model long-range dependencies in sequential data is fundamental to many machine learning applications, including language modeling, time-series forecasting, and multimodal reasoning. 
Classical architectures such as recurrent neural networks (RNNs) and transformers have achieved remarkable performance in these tasks. 
However, both approaches show intrinsic limitations as the sequence length increases. RNN-based models suffer from vanishing or exploding gradients that degrade their ability to preserve long-term information~\cite{hochreiter1997long,bengio1994learning}.
Transformers~\cite{vaswani2017attention} address this issue using attention mechanisms, but their quadratic computational complexity with respect to sequence length.
A number of recent works attempt to mitigate this limitation through efficient or long-sequence attention variants, including sparse attention \cite{child2019generating}, low-rank and kernelized approximations \cite{katharopoulos2020transformers,choromanski2020rethinking}, and state-space-inspired architectures such as linear attention and hybrid models \cite{tay2022efficient,dao2022flashattention}.
More recently, State-Space Models (SSMs) \cite{gu2021efficiently,gu2021combining} have emerged as an efficient alternative, enabling linear-time sequence processing through structured state transitions. 
Despite these advances, existing approaches remain fundamentally classical, relying on fixed-dimensional vector representations of memory.



Quantum information theory offers a fundamentally different paradigm for representing and evolving information. In quantum systems, information is encoded in the amplitudes of vectors in a complex Hilbert space. A quantum state can represent a coherent superposition of multiple basis states, allowing a compact representation of information that scales exponentially with the number of qubits~\cite{nielsen2010quantum,preskill2018quantum}. Moreover, quantum evolution is governed by unitary operators, which preserve the norm of the quantum state and therefore avoid the exponential growth or decay typically encountered in classical dynamical systems.


Far apart from prior studies on attention-based long-sequence modeling methods, this work explores a different perspective on the problem mentioned: \textit{\textbf{Can memory itself be fundamentally restructured using quantum computation?}} 
Motivated by the key properties in quantum computing, i.e., superposition and unitary evolution, we introduce Quantum Long-Attention Memory (QLAM), a hybrid quantum-classical memory framework where contextual information is stored and processed directly on a quantum device. QLAM leverages actual quantum states and operations to represent and manipulate memory. 
At each time step, the memory is encoded as a quantum state that evolves through parameterized quantum circuits conditioned on the input, while task-relevant information is retrieved via query-dependent measurements.

Compared to classical long-attention methods, QLAM provides a more suitable memory abstraction for long-sequence modeling because it does not require storing or explicitly comparing all historical token representations.
Classical long-attention mechanisms typically improve scalability by sparsifying, compressing, or approximating the attention matrix, but they still rely on classical key-value memories whose capacity grows with sequence length or is constrained by fixed-dimensional compression. 
In particular, sparse attention methods \cite{child2019generating,zaheer2020big} reduce complexity at the cost of potentially missing global dependencies, while low-rank and kernel-based approximations \cite{katharopoulos2020transformers,choromanski2020rethinking} introduce approximation errors that can degrade representation fidelity for complex interactions. 
Linear attention and SSMs methods \cite{tay2022efficient,dao2022flashattention,gu2021efficiently} achieve favorable scaling but often compress historical information into fixed-size states, which may limit expressivity and lead to information loss over long horizons.
In contrast, QLAM encodes contextual information in a quantum state, in which multiple memory components can be represented in superposition and evolved jointly by unitary transformations. 
This allows long-range information to be maintained in a compact and globally integrated memory representation, while query-dependent measurements retrieve task-relevant information without explicitly computing all pairwise token interactions. 


This formulation departs from conventional attention and memory mechanisms in several important ways. First, instead of storing context as a collection of token-wise embeddings or key-value pairs, QLAM represents memory as a global quantum state, enabling a highly expressive and distributed representation of information.
Second, memory updates are controlled by unitary transformations, which inherently preserve the state's structure and provide a stable mechanism for information propagation. Third, retrieval is performed via measurement, enabling query-dependent information extraction without explicitly computing pairwise interactions between tokens.
Together, these properties suggest a fundamentally different paradigm for sequence modeling, in which memory is no longer a passive storage but an actively evolving quantum system.
Finally, QLAM is presented as a new design for hybrid models that integrate quantum computation into the core of memory representation.




\textbf{Contributions of this Work.} 
This work presents a new Quantum Long-Attention Memory (QLAM) method, a hybrid classical-quantum memory mechanism where memory is represented and updated using quantum states and circuits. 
First, we propose a query-dependent measurement-based retrieval mechanism that generalizes attention to the quantum setting. 
Second, we conduct an empirical exploration to evaluate the feasibility and potential of quantum memory in practical learning settings. 
Finally, our results suggest that hybrid quantum memory is a promising direction for rethinking sequence modeling, opening new opportunities at the intersection of machine learning and quantum computation.

\section{Related Work}

\subsection{Sequential Token Modeling}

Early work for processing sequential data relied on state-tracking architectures, most notably recurrent neural networks (RNNs) \cite{rumelhart1985learning,elman1990finding,hochreiter1997long,cho2014learning}. 
This approach continuously updated a compressed hidden state, and, in doing so, these networks, as well as gated variants such as long short-term memory networks (LSTMs) and gated recurrent units (GRUs), excelled at localized sequence modeling. However, these models exhibited performance degradation over longer sequences. The introduction of Transformers \cite{vaswani2017attention} fundamentally shifted this landscape by enabling global self-attention, which allowed direct, uncompressed routing between arbitrary sequence positions. 
This paradigm has driven breakthroughs in natural language processing \cite{devlin2019bert,brown2020language,radford2019language} and computer vision \cite{dosovitskiy2020image,liu2021swin}. 
However, this expressivity comes at a high computational cost due to the quadratic complexity of attention with respect to sequence length, motivating research on efficient attention \cite{katharopoulos2020transformers,choromanski2020rethinking}. 
Recent literature has turned to state-space models (SSMs) \cite{gu2021efficiently,smith2022simplified,gu2023mamba} to bypass the self-attention bottleneck. 
By formalizing the sequence processing through structured linear dynamical systems, architectures such as S4 \cite{gu2021efficiently}, S5 \cite{smith2022simplified}, and Mamba \cite{gu2023mamba} achieve linear-time scaling without sacrificing the modeling fidelity of Transformers for long-context tasks.

\begin{figure*}[t]
    \centering
    \includegraphics[width=0.9\linewidth]{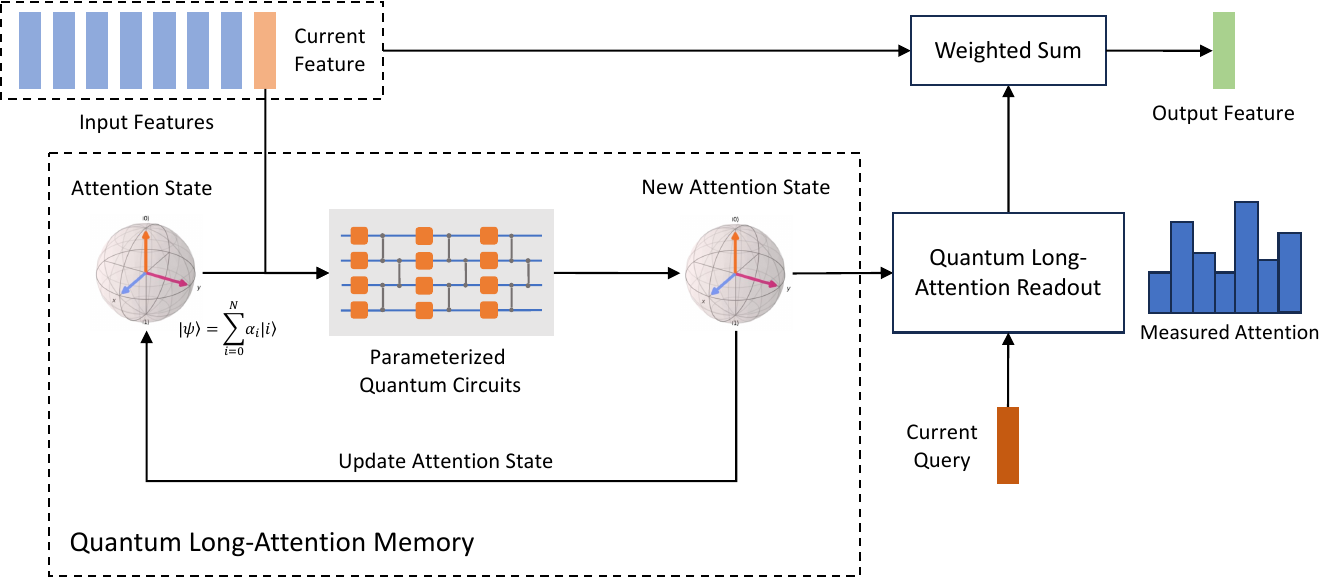}
    \caption{Overview framework of the proposed Quantum Long-Attention Memory.}
    \label{fig:overview_framework}
\end{figure*}

\subsection{Quantum Machine Learning}

Quantum machine learning (QML) has emerged as a paradigm to augment machine learning by exploiting quantum mechanics for enhanced representational capacity and computational efficiency~\cite{biamonte2017quantum,schuld2015introduction,schuld2019quantum}. 
Early developments in QML primarily focused on accelerating classical linear machine learning algorithms through quantum speedups. 
These works demonstrated quantum advantages in tasks such as clustering~\cite{lloyd2013quantum,nguyen2023quantum,nguyen2024qclusformer}, principal component analysis~\cite{lloyd2014quantum}, least-squares fitting~\cite{schuld2016prediction,kerenidis2020quantum}, and binary classification~\cite{rebentrost2014quantum}. 
Within the constraints of the noisy intermediate-scale quantum (NISQ) era, variational quantum circuits (VQCs) \cite{cerezo2021variational}, or parameterized quantum circuits \cite{benedetti2019parameterized}, serve as the foundational architecture. 
These models have achieved notable traction across diverse domains, including standard classification \cite{schuld2020circuit,nguyen2024hierarchical,nguyen2025qmoe}, optimization \cite{farhi2014quantum,zhou2020quantum,holliday2025quadro}, generative modeling \cite{benedetti2019generative,huang2021experimental,nguyen2025diffusion}, and reinforcement learning \cite{chen2022variational}. 
Building upon this framework, several architectures have been proposed to extend classical deep learning concepts into the quantum domain. 
For instance, quantum convolutional neural networks~\cite{cong2019quantum} adapt convolutional structures with reduced parameterization, while quantum autoencoders~\cite{romero2017quantum} enable compression and representation learning of quantum states. 
Additionally, quantum neural networks based on PQCs~\cite{panella2011neural,mitarai2018quantum,nguyen2024quantum,nguyen2024quantum_brain} provide a flexible foundation for learning nonlinear mappings in hybrid settings.
Beyond developments in QML, recent studies have also explored physics-aware machine learning for quantum material discovery, particularly in the analysis of two-dimensional quantum materials from optical microscopy images \cite{nguyen2025phi_adapt,pandey2026openqlaw,nguyen2026qupaint}. 
However, the limitations of these VQC formulations become apparent when the focus shifts to dynamic, time-series data. 
In particular, embedding complex temporal dependencies into shallow, parameterized unitary operations presents a blocking point. 
As such, recent work proposes quantum architectures tailored for sequential data. 
Approaches such as quantum recurrent neural networks and quantum dynamical systems \cite{beer2020training,li2023quantum} aim to model temporal evolution via continuous quantum-state transitions.
In contrast to prior approaches that focus on either recurrent dynamics, our work explores a complementary direction by leveraging quantum superposition as a memory mechanism for sequence modeling. 
The proposed Quantum Long-Attention Memory (QLAM) encodes the evolving sequence context into a structured quantum state, allowing information from multiple tokens to be aggregated implicitly through superposition. 
This enables attention-like global interactions while maintaining a compact and recurrently updated memory representation, offering a novel perspective for long-sequence modeling in QML.

\section{Background}

\subsection{Transformers Attention Mechanisms}

Attention mechanisms \cite{vaswani2017attention} enable models to selectively focus on relevant parts of an input sequence. In transformer architectures, self-attention computes pairwise interactions between tokens through a query-key-value formulation.
Given queries $Q=\{q_i \in \mathbb{R}^d\}_{i=1}^T$, keys $K=\{k_i \in \mathbb{R}^d\}_{i=1}^T$, and values $V=\{v_i \in \mathbb{R}^d\}_{i=1}^T$, the attention output is computed as follows,
\begin{equation}
    \text{Attention}(Q, K, V) = \text{softmax}\left( \frac{Q K^\top}{\sqrt{d}} \right) V.
\end{equation}
This formulation allows each token to attend to all other tokens in the sequence, enabling flexible modeling of long-range dependencies. 
However, the computational complexity of self-attention scales quadratically with the sequence length, which becomes prohibitive for very long contexts.

Several methods have been proposed to reduce this cost, including linear attention mechanisms \cite{katharopoulos2020transformers,choromanski2020rethinking} and state-space models \cite{gu2021efficiently,gu2021combining,gu2023mamba} that implicitly capture contextual information through recurrent state updates. In contrast to explicit pairwise interactions, these methods rely on compressed representations of the sequence history.

\subsection{State-Space Models (SSMs)}

State-space models (SSMs) describe the evolution of a latent dynamical system driven by an input sequence. In discrete time, a linear state-space model is typically expressed as:
\begin{equation}
\begin{split}
    h_{t+1} &= A h_t + B x_t \\
    y_t &= C h_t,
\end{split}
\end{equation}
where $x_t \in \mathbb{R}^d$ is the input, $h_t \in \mathbb{R}^n$ is the latent state, and $y_t$ denotes the output.
The matrices $A$, $B$, and $C$ determine the state transition, input injection, and readout operators, respectively.

SSMs have recently gained attention in machine learning due to their ability to model long sequences with linear computational complexity. Modern architectures based on structured SSMs introduce parameterizations that enable efficient convolutional implementations and stable training dynamics. In these models, the latent state $h_t$ serves as a compressed representation of the historical sequence, allowing information from earlier tokens to influence future predictions.

Despite these advantages, classical SSMs rely on additive accumulation of information within a finite-dimensional state vector. As the sequence length grows, the model must continuously compress historical information into a bounded representation. 
This limitation may reduce the expressive capacity of the memory mechanism and can lead to information loss over long time horizons.

\begin{figure*}[t]
    \centering
    \includegraphics[width=0.7\linewidth]{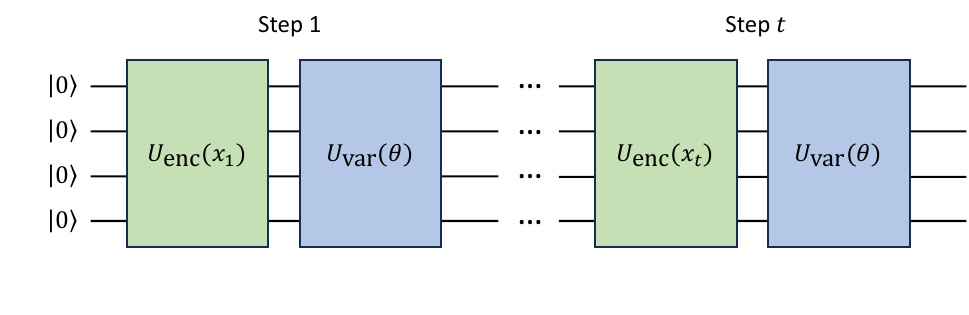}
    \vspace{-8mm}
    \caption{The quantum long-attention memory is evolved across input features.}
    \label{fig:memory_evolution}
\end{figure*}

\section{Quantum Long-Attention Memory (QLAM)}


In this section, we introduce Quantum Long-Attention Memory (QLAM), a quantum-based memory mechanism designed for long-attention-sequence modeling, as presented in Figure \ref{fig:overview_framework}.
The key idea is to represent the memory state as a vector in a complex Hilbert space and evolve it through unitary transformations conditioned on the input sequence.
This formulation enables stable memory propagation and allows contextual information to be stored through superposition.

\subsection{Problem Formulation}

We consider a sequential prediction problem over an input sequence:
\begin{equation}
    \{x_1, x_2, \dots, x_T\}, \quad x_t \in \mathcal{X},
\end{equation}
where $\mathcal{X}$ denotes the input space, e.g., tokens, image patches, or sensor measurements.
The objective is to learn a model that produces outputs:
\begin{equation}
    \{y_1, y_2, \dots, y_T\}, \quad y \in \mathcal{Y},
\end{equation}
such that each output $y_t$ depends on the entire history:
\begin{equation}
    y_t = f(x_1, x_2, \dots, x_t).
\end{equation}
Most classical approaches maintain a hidden state $h_t \in \mathbb{R}^d$ or a key-value memory.
In contrast, QLAM maintains a quantum memory state that evolves over time steps.

\subsection{Quantum Long-Memory Representation}

Given an input sequence $\{x_t\}_{t=1}^T$, standard attention computes:
\begin{equation}
    y_t = \sum_{s \leq t} \alpha_{t,s} v_s, 
    \quad \alpha_{t,s} \propto \exp(q_t^\top k_s)
\end{equation}
which requires evaluating all pairwise token interactions.
In contrast, our approach replaces this computation with quantum-state accumulation, and attention weights are reconstructed from this state.

Let the memory state at time $t$ be represented by a normalized vector in a complex Hilbert space:
\begin{equation}
    |\psi_t\rangle \in \mathbb{C}^{2^n}, \quad \langle \psi_t | \psi_t \rangle = 1,
\end{equation}
where $n$ denotes the number of qubits used to represent the memory.
The quantum state can be expressed in the computational basis as:
\begin{equation}
    |\psi_t\rangle = \sum_{i=1}^{2^n-1} \alpha_t^{(i)} |i\rangle,
\end{equation}
where $\alpha_t^{(i)} \in \mathbb{C}$ are complex amplitudes, encoding information about the sequence.
Each basis state $|i\rangle$ corresponds to an element in a memory configuration.
Then, the full state represents a superposition of the memory.
This allows the memory state to encode a coherent superposition of multiple contextual signals and to evolve over timesteps.
Compared to classical approaches where the classical states $h_t \in \mathbb{R}^d$ require the memory complexity of $O(d)$, the representation of QLAM for the state $|\psi_t\rangle \in \mathbb{C}^{2^n}$ require the memory complexity of $O(n) \sim O(\log_2d)$, which makes the QLAM more efficient in memory storage.


\subsection{Quantum Long-Memory Evolution}

The memory state evolves through unitary transformations conditioned on the input token $x_t$.
Formally, the memory update rule is defined as:
\begin{equation}
    |\psi_t\rangle = U(x_t, \theta) | \psi_{t-1}\rangle,
\end{equation}
where $U(x_t, \theta)$ is a parameterized unitary operator.
Formally, we decompose the unitary as:
\begin{equation}
    U(x_t, \theta) = U_\text{var}(\theta) U_\text{enc}(x_t),
\end{equation}
where $U_\text{enc}(x_t)$ is an input encoding operator and $U_\text{var}(\theta)$ is a trainable parameterized circuit.
This separation clarifies the roles of data injection and learned transformation.
Initially, the quantum long-attention memory is defined as $|\psi_1\rangle = |0\rangle^{\otimes n}$, showing the attention is focused on the first feature $x_1$.
Then, the attention memory is evolved through input features $x_t$, as shown in Figure \ref{fig:memory_evolution}.

A key property of this update rule is that unitary transformations satisfy:
\begin{equation}
    U(x_t, \theta)^\dagger U(x_t, \theta) = I,
\end{equation}
which implies:
\begin{equation}
    \| |\psi_t\rangle \| = \| |\psi_{t-1}\rangle \|.
\end{equation}
Therefore, the memory dynamics are norm-preserving, ensuring stable propagation across long sequences. 
This property contrasts with classical recurrent systems in which repeated matrix multiplications may lead to exploding or vanishing states.

\subsection{Quantum Long-Attention Readout}

\begin{figure*}[t]
    \centering
    \includegraphics[width=\linewidth]{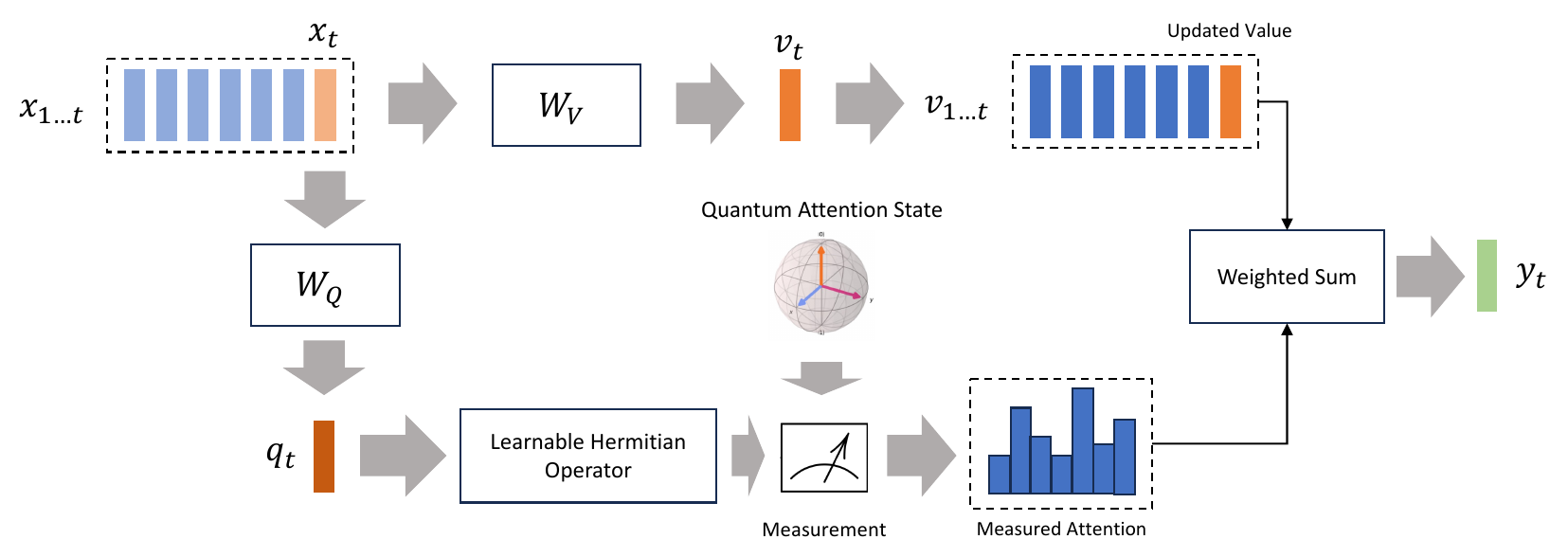}
    \vspace{-8mm}
    \caption{Computation flow of quantum long-attention readout.}
    \label{fig:quantum_attention_readout}
\end{figure*}

\begin{figure}[t]
    \centering
    \includegraphics[width=\linewidth]{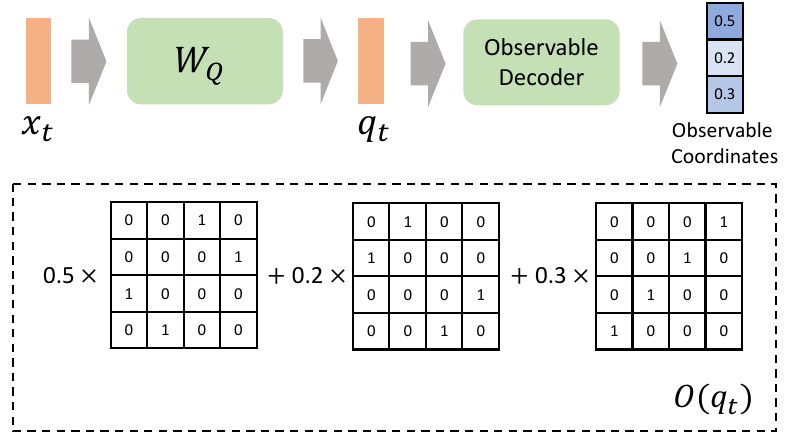}
    \vspace{-4mm}
    \caption{The observable $O(q_t)$ is formed as a weighted combination of multiple Pauli matrices based on the query $q_t$.}
    \label{fig:observable_computing}
\end{figure}

To extract task-relevant information from the quantum memory state, we introduce a measurement-based readout mechanism that serves as an attention query $q_t = W_Q x_t$, as shown in Figure \ref{fig:quantum_attention_readout}.
Unlike classical attention, which retrieves information through explicit similarity computations over stored representations, QLAM performs retrieval by probing the quantum state with a query-conditioned observable.
For each previous token $s < t$, we compute an attention through a measurement operator:
\begin{equation}
    \alpha_{t,s} = \langle \psi_t | O(q_t) | \psi_t \rangle,
\end{equation}
where $O(q_t) \in \mathbb{C}^{2^n \times 2^n}$ is a learnable Hermitian observable operator parameterized by the query $q_t$.
In detail, given a set of Pauli matrices $\mathcal{P} = \{O_i\}_{i=1}^p$, we project the query $q_t$ via a small neural network as a learnable observable decoder into observable coordinates $\{\gamma_i \in \mathbb{R}\}_{i=1}^p$ indicating the weighted combination of the Pauli matrices:
\begin{equation}
    O(q_t) = \sum_{i=1}^p \gamma_i O_i.
\end{equation}
Since $\gamma_i$ is real, $\gamma_i^* = \gamma_i$, and $O_i$ is Hermitian, $O_i^\dagger = O_i$, $O(q_t)$ is guaranteed to be Hermitian:
\begin{equation}
\begin{split}
    O(q_t)^\dagger &= \left(\sum_{i=1}^p \gamma_i O_i\right)^\dagger
    = \sum_{i=1}^p \gamma_i^* O_i^\dagger \\
    &= \sum_{i=1}^p \gamma_i O_i 
    = O(q_t).
\end{split}
\end{equation}
This construction allows the query to select an adaptive measurement basis while keeping the measurement process physically valid.
The formulation defines a query-conditioned projection of the quantum memory, where the observable determines how information is extracted from the state, as illustrated in Figure \ref{fig:observable_computing}.

In practice, expectation values are approximated through repeated measurement:
\begin{equation}
\begin{split}
    o^{(1)}_{t,s},\dots, o^{(m)}_{t,s} &\sim \mathcal{M}(O(q_t), |\psi_t\rangle) \\
    \alpha_{t,s} &= \frac{1}{m} \sum_{i=1}^m o^{(i)}_{t,s}.
\end{split}
\end{equation}
The estimator is unbiased and converges to the exact expectation as $m$ is increased.
This introduces a controllable stochastic component that can improve robustness.

\begin{figure*}[t]
    \centering
    \includegraphics[width=\linewidth]{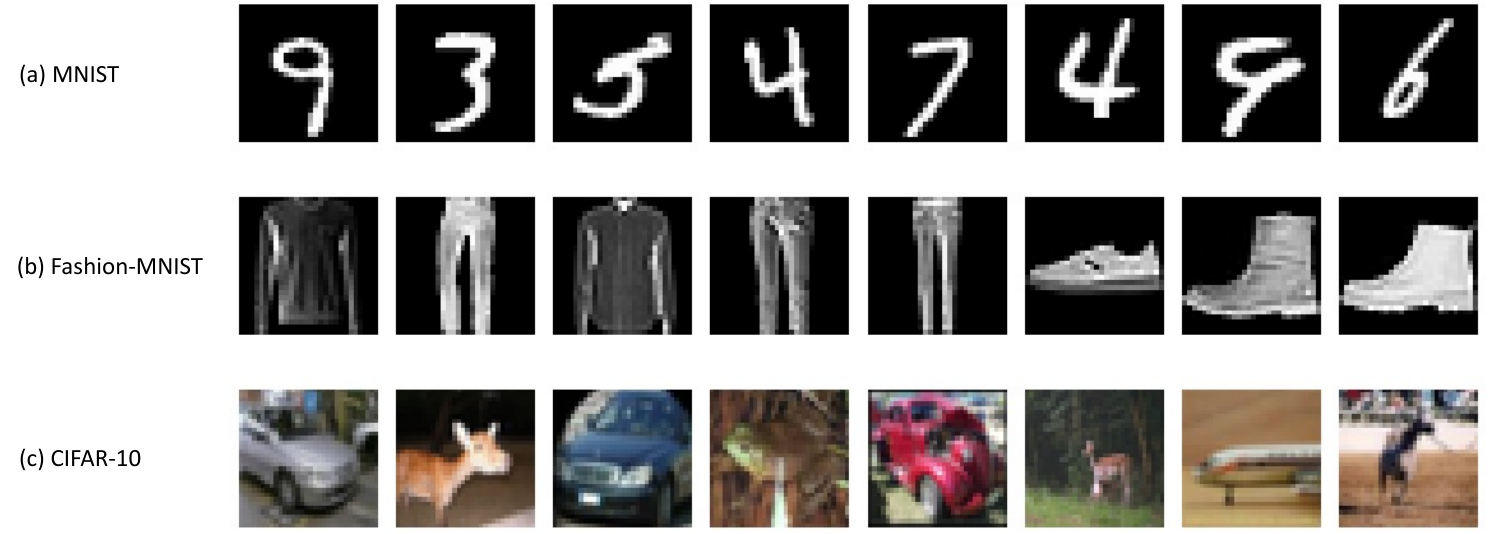}
    \vspace{-4mm}
    \caption{
    Visualization of sample inputs from the benchmark datasets used in our experiments, including (a) MNIST~\cite{lecun2010mnist}, (b) Fashion-MNIST~\cite{xiao2017fashion}, and (c) CIFAR-10~\cite{krizhevsky2009learning}.
    These datasets are later reformulated as sequential inputs by flattening each image into a one-dimensional token sequence, enabling evaluation of sequence modeling capabilities.
    }
    \label{fig:input_samples}
\end{figure*}

\begin{figure*}[t]
    \centering
    \includegraphics[width=\linewidth]{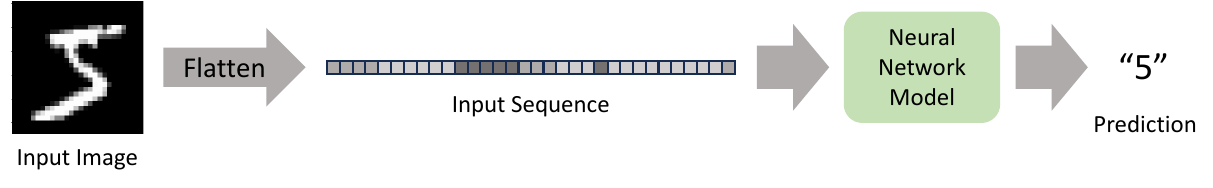}
    \vspace{-8mm}
    \caption{Evaluation protocol for sequential modeling with images.}
    \label{fig:input_computing}
\end{figure*}

\section{Experimental Results}

\subsection{Experimental Setup}

We evaluate the proposed Quantum Long-Attention Memory framework on image classification. We evaluate the proposed QLAM framework on image classification tasks reformulated as sequence modeling problems, a standard protocol for assessing long-range dependency modeling.
Specifically, we consider MNIST~\cite{lecun2010mnist}, Fashion-MNIST~\cite{xiao2017fashion}, and CIFAR-10~\cite{krizhevsky2009learning}, as shown in Figure \ref{fig:input_samples}, where each image is converted into a one-dimensional sequence and processed token-by-token.
This formulation removes spatial inductive biases and forces models to rely entirely on sequential reasoning, making it particularly suitable for evaluating memory mechanisms.
For convenience, we denote the resulting datasets as \textit{sMNIST}, \textit{sFashion-MNIST}, and \textit{sCIFAR-10}, respectively.

For MNIST and Fashion-MNIST, each $28 \times 28$ grayscale image is reshaped into a sequence of length $784$, where each token corresponds to a single pixel intensity.
For CIFAR-10, each $32 \times 32 \times 3$ image is flattened into a sequence of length $3072$, with RGB channels concatenated along the sequence dimension.
All pixel values are normalized to $[0,1]$ and treated as continuous inputs.
The model processes the sequence causally, i.e., tokens are consumed sequentially without access to future information, thereby mimicking standard autoregressive sequence modeling.
A final classification head is applied on the last hidden state to predict the image label.
The inference process is illustrated in Figure \ref{fig:input_computing}.

We compare QLAM against representative baselines spanning different sequence modeling paradigms: 
(i) a recurrent neural network (RNN)~\cite{rumelhart1985learning}, which relies on iterative hidden state updates,
(ii) a Transformer encoder~\cite{vaswani2017attention}, which models global dependencies via self-attention, and 
(iii) a state-space model (SSM)~\cite{gu2021efficiently}, which captures long-range interactions through structured linear dynamics.
All models are implemented with comparable parameter budgets to ensure a fair comparison, and share similar input projections and output classifiers.

To obtain statistically reliable results, we adopt a 10-fold evaluation protocol.
For each dataset, we train and evaluate the models across 10 independent runs with different data splits.
We report both single-run performance and aggregated statistics, i.e., mean and standard deviation, providing a comprehensive assessment of both accuracy and stability.


\begin{table*}[t]
\centering
\caption{Experimental accuracies (\%) of the proposed approach compared to prior baselines.}
\begin{tabular}{l|cccccccccc|c}
\Xhline{2\arrayrulewidth}
\textbf{Method} & \textbf{Fold-1} & \textbf{Fold-2} & \textbf{Fold-3} & \textbf{Fold-4} & \textbf{Fold-5} & \textbf{Fold-6} & \textbf{Fold-7} & \textbf{Fold-8} & \textbf{Fold-9} & \textbf{Fold-10} & \textbf{Mean $\pm$ Std} \\
\Xhline{2\arrayrulewidth}
\multicolumn{12}{c}{\textbf{sMNIST}~\cite{lecun2010mnist}} \\
\hline
RNN~\cite{rumelhart1985learning} & 11.2 & 11.5 & 11.6 & 11.3 & 11.4 & 11.5 & 11.3 & 11.6 & 11.4 & 11.5 & 11.4 $\pm$ 0.13 \\
Transformer~\cite{vaswani2017attention} & 91.0 & 91.5 & 91.2 & 91.4 & 91.3 & 91.6 & 91.2 & 91.4 & 91.3 & 91.5 & 91.3 $\pm$ 0.18 \\
SSM~\cite{gu2021efficiently} & 90.9 & 91.3 & 91.1 & 91.2 & 91.0 & 91.4 & 91.2 & 91.1 & 91.3 & 91.2 & 91.2 $\pm$ 0.15 \\
\hline
\textbf{QAM} & \textbf{92.3} & \textbf{92.7} & \textbf{92.6} & \textbf{92.5} & \textbf{92.4} & \textbf{92.8} & \textbf{92.6} & \textbf{92.5} & \textbf{92.7} & \textbf{92.6} & \textbf{92.6 $\pm$ 0.15} \\

\Xhline{2\arrayrulewidth}
\multicolumn{12}{c}{\textbf{sFashion-MNIST}~\cite{xiao2017fashion}} \\
\hline
RNN~\cite{rumelhart1985learning} & 12.5 & 12.8 & 12.7 & 12.6 & 12.7 & 12.9 & 12.6 & 12.8 & 12.7 & 12.8 & 12.7 $\pm$ 0.12 \\
Transformer~\cite{vaswani2017attention} & 79.8 & 80.4 & 80.1 & 80.3 & 80.2 & 80.5 & 80.0 & 80.3 & 80.2 & 80.4 & 80.2 $\pm$ 0.21 \\
SSM~\cite{gu2021efficiently} & 79.2 & 79.8 & 79.5 & 79.7 & 79.6 & 79.9 & 79.4 & 79.7 & 79.5 & 79.8 & 79.6 $\pm$ 0.22 \\
\hline
\textbf{QAM} & \textbf{81.1} & \textbf{81.6} & \textbf{81.5} & \textbf{81.3} & \textbf{81.4} & \textbf{81.7} & \textbf{81.2} & \textbf{81.5} & \textbf{81.4} & \textbf{81.6} & \textbf{81.4 $\pm$ 0.19} \\

\Xhline{2\arrayrulewidth}
\multicolumn{12}{c}{\textbf{sCIFAR-10}~\cite{krizhevsky2009learning}} \\
\hline
RNN~\cite{rumelhart1985learning} & 13.5 & 13.8 & 13.6 & 13.7 & 13.7 & 13.9 & 13.6 & 13.8 & 13.7 & 13.8 & 13.7 $\pm$ 0.13 \\
Transformer~\cite{vaswani2017attention} & 52.4 & 53.2 & 52.8 & 53.0 & 52.9 & 53.3 & 52.7 & 53.1 & 52.9 & 53.2 & 53.0 $\pm$ 0.28 \\
SSM~\cite{gu2021efficiently} & 51.3 & 52.0 & 51.6 & 51.8 & 51.7 & 52.1 & 51.5 & 51.9 & 51.7 & 52.0 & 51.8 $\pm$ 0.25 \\
\hline
\textbf{QAM} & \textbf{53.1} & \textbf{53.8} & \textbf{53.6} & \textbf{53.4} & \textbf{53.5} & \textbf{53.9} & \textbf{53.3} & \textbf{53.7} & \textbf{53.5} & \textbf{53.8} & \textbf{53.6 $\pm$ 0.26} \\

\Xhline{2\arrayrulewidth}
\end{tabular}
\label{tab:exp_results}
\end{table*}

\subsection{Implementation Details}

The models are implemented in PyTorch \cite{paszke2019pytorch}, with the quantum components of QLAM implemented using a hybrid quantum-classical simulation framework with the PennyLane library \cite{bergholm2018pennylane}.
The input at each timestep is first processed by a lightweight classical encoder, consisting of a linear projection layer that maps scalar pixel values into a higher-dimensional embedding.
This embedding is then used to parameterize the quantum circuit through angle encoding.
The initial quantum state is initialized to the computational basis state $|0\rangle^{\otimes n}$.
All models are trained using the Adam optimizer\cite{kingma2014adam} with an initial learning rate of $10^{-3}$.
We train for 30 epochs on sMNIST and sFashion-MNIST, and 50 epochs on sCIFAR-10.
A cosine learning rate scheduler is applied.
The batch size is set to 128 for all experiments.
For baseline models, we use comparable model sizes to ensure a fair comparison.
We evaluate model performance using standard classification accuracy on the test set.

\subsection{Evaluation Results on sMNIST}

As shown on the sMNIST benchmark in Table~\ref{tab:exp_results}, all sequence-based models significantly outperform the vanilla RNN, confirming the inherent difficulty of modeling long-range dependencies with purely recurrent dynamics.
The RNN fails to retain information across long pixel sequences, resulting in near-random performance.
In contrast, both the Transformer~\cite{vaswani2017attention} and SSM~\cite{gu2021efficiently} achieve strong performance around $91\%$, demonstrating their ability to capture long-range interactions through attention mechanisms and structured state transitions, respectively.

Our proposed QLAM further improves accuracy to $92.6 \pm 0.15$ in the 10-fold evaluation, consistently outperforming all baselines across all folds.
Notably, the improvement is achieved without increasing model complexity, indicating that the gain arises from a more effective memory mechanism rather than scale.
The low standard deviation highlights the stability of the proposed approach, suggesting that QLAM mitigates optimization issues commonly observed in long-sequence modeling.
From a modeling perspective, this result supports the hypothesis that superposition-based memory aggregation enables a richer representation of global dependencies than classical additive accumulation.

\subsection{Evaluation Results on sFashion-MNIST}

The sFashion-MNIST dataset is more challenging due to higher intra-class variability and more complex visual structures, requiring models to capture both fine-grained and global dependencies.
As reported in Table~\ref{tab:exp_results}, the Transformer and SSM achieve $80.2\%$ and $79.6\%$, respectively, indicating a degradation compared to sMNIST due to increased task complexity.

In contrast, QLAM achieves $81.4 \pm 0.19$ under the 10-fold protocol, consistently outperforming both baselines.
The performance gap is more noticeable than on sMNIST, suggesting that QLAM provides greater benefits when the underlying data distribution is more complex.
This behavior can be attributed to QLAM's ability to encode multiple dependency patterns simultaneously via superposition, enabling the model to better capture diverse visual features.

Furthermore, QLAM exhibits lower variance across folds than baseline models, indicating greater robustness to data splits and initialization.
This stability is particularly important in more complex datasets, where optimization landscapes are typically more irregular.
Overall, these results demonstrate that QLAM not only improves accuracy but also enhances training stability in challenging sequential learning scenarios.

\subsection{Evaluation Results on sCIFAR-10}

The sCIFAR-10 benchmark presents a substantially more difficult setting due to higher-dimensional inputs, color channels, and richer semantic content.
All models experience a noticeable drop in performance compared to grayscale datasets, reflecting the increased difficulty of modeling long-range dependencies in high-entropy sequences.

As shown in Table~\ref{tab:exp_results}, the Transformer and SSM achieve $52.4\%$ and $51.3\%$, respectively, while QLAM improves the performance to $53.1\%$.
Under 10-fold evaluation, QLAM achieves $53.6 \pm 0.26$, consistently outperforming both baselines across all folds.
Although the absolute improvement is smaller compared to sMNIST and sFashion-MNIST, the gain remains consistent, which is non-trivial given the increased complexity of the task.

Importantly, the variance of QLAM remains competitive despite the greater difficulty, indicating that the proposed memory mechanism scales favorably to more complex settings.
This suggests that QLAM provides a more expressive yet stable framework for aggregating long-range information, even when the input distribution becomes significantly more diverse.
From a broader perspective, these results highlight that the advantage of QLAM is not limited to simple datasets, but extends to more realistic and challenging scenarios where both expressivity and stability are required.

\section{Conclusions}


In this work, we have introduced Quantum Long-Attention Memory (QLAM), a novel sequence modeling framework that extends state-space models by representing memory as a quantum state and evolving it through unitary transformations. 
This formulation has addressed key limitations of existing approaches, where transformers suffer from quadratic complexity and classical state-space models rely on additive updates that may restrict expressivity in long sequences. 
By encoding historical information in a superposition-based quantum state, QLAM has provided a compact global memory representation, enabling implicit capture of interactions across tokens without explicit pairwise computations. 
The use of unitary dynamics has ensured stable propagation of information over long horizons, while the measurement-based readout has provided a query-dependent mechanism for retrieving task-relevant information, effectively generalizing attention within this framework. 
Empirical results on sMNIST, sFashion-MNIST, and sCIFAR-10 have demonstrated that QLAM consistently outperforms recurrent, transformer-based, and state-space baselines under comparable settings, highlighting improvements in both accuracy and stability for long-range sequence modeling. 
While these findings validate the potential of quantum-state-based memory, further work is needed to further study the theoretical understanding of its properties, explore implementations on quantum hardware, and extend the framework to large-scale language and multimodal tasks. Overall, QLAM offers a new perspective on sequence modeling by fundamentally rethinking memory through quantum computation.

{
    \small
    \bibliographystyle{IEEEtran}
    \bibliography{IEEE_main}

@String{Computing = "Computing" }

@String{Computer = "{IEEE} Computer" }

@String{Springer = "Springer-Verlag" }

@article{preskill2018quantum,
  title={Quantum computing in the NISQ era and beyond},
  author={Preskill, John},
  journal={Quantum},
  volume={2},
  pages={79},
  year={2018},
  publisher={Verein zur F{\"o}rderung des Open Access Publizierens in den Quantenwissenschaften}
}

@article{lecun2010mnist,
  title={MNIST handwritten digit database},
  author={LeCun, Yann and Cortes, Corinna and Burges, CJ},
  journal={ATT Labs [Online]. Available: http://yann.lecun.com/exdb/mnist},
  volume={2},
  year={2010}
}

@article{xiao2017fashion,
  title={Fashion-mnist: a novel image dataset for benchmarking machine learning algorithms},
  author={Xiao, Han and Rasul, Kashif and Vollgraf, Roland},
  journal={arXiv preprint arXiv:1708.07747},
  year={2017}
}

@article{krizhevsky2009learning,
  title={Learning multiple layers of features from tiny images},
  author={Krizhevsky, Alex and Hinton, Geoffrey and others},
  year={2009},
  publisher={Toronto, ON, Canada}
}

@article{vaswani2017attention,
  title={Attention is all you need},
  author={Vaswani, Ashish and Shazeer, Noam and Parmar, Niki and Uszkoreit, Jakob and Jones, Llion and Gomez, Aidan N and Kaiser, {\L}ukasz and Polosukhin, Illia},
  journal={Advances in neural information processing systems},
  volume={30},
  year={2017}
}

@article{gu2021efficiently,
  title={Efficiently modeling long sequences with structured state spaces},
  author={Gu, Albert and Goel, Karan and R{\'e}, Christopher},
  journal={arXiv preprint arXiv:2111.00396},
  year={2021}
}

@techreport{rumelhart1985learning,
  title={Learning internal representations by error propagation},
  author={Rumelhart, David E and Hinton, Geoffrey E and Williams, Ronald J},
  year={1985}
}

@article{paszke2019pytorch,
  title={Pytorch: An imperative style, high-performance deep learning library},
  author={Paszke, Adam and Gross, Sam and Massa, Francisco and Lerer, Adam and Bradbury, James and Chanan, Gregory and Killeen, Trevor and Lin, Zeming and Gimelshein, Natalia and Antiga, Luca and others},
  journal={Advances in neural information processing systems},
  volume={32},
  year={2019}
}

@article{bergholm2018pennylane,
  title={Pennylane: Automatic differentiation of hybrid quantum-classical computations},
  author={Bergholm, Ville and Izaac, Josh and Schuld, Maria and Gogolin, Christian and Ahmed, Shahnawaz and Ajith, Vishnu and Alam, M Sohaib and Alonso-Linaje, Guillermo and AkashNarayanan, Bharath and Asadi, Ali and others},
  journal={arXiv preprint arXiv:1811.04968},
  year={2018}
}

@article{kingma2014adam,
  title={Adam: A method for stochastic optimization},
  author={Kingma, Diederik P and Ba, Jimmy},
  journal={arXiv preprint arXiv:1412.6980},
  year={2014}
}

@article{hochreiter1997long,
  title={Long short-term memory},
  author={Hochreiter, Sepp and Schmidhuber, J{\"u}rgen},
  journal={Neural computation},
  volume={9},
  number={8},
  pages={1735--1780},
  year={1997},
  publisher={MIT press}
}

@article{bengio1994learning,
  title={Learning long-term dependencies with gradient descent is difficult},
  author={Bengio, Yoshua and Simard, Patrice and Frasconi, Paolo},
  journal={IEEE transactions on neural networks},
  volume={5},
  number={2},
  pages={157--166},
  year={1994},
  publisher={IEEE}
}

@inproceedings{cho2014learning,
  title={Learning phrase representations using RNN encoder--decoder for statistical machine translation},
  author={Cho, Kyunghyun and Van Merri{\"e}nboer, Bart and Gul{\c{c}}ehre, {\c{C}}a{\u{g}}lar and Bahdanau, Dzmitry and Bougares, Fethi and Schwenk, Holger and Bengio, Yoshua},
  booktitle={Proceedings of the 2014 conference on empirical methods in natural language processing (EMNLP)},
  pages={1724--1734},
  year={2014}
}

@article{gu2023mamba,
  title={Mamba: Linear-time sequence modeling with selective state spaces},
  author={Gu, Albert and Dao, Tri},
  journal={arXiv preprint arXiv:2312.00752},
  year={2023}
}

@article{elman1990finding,
  title={Finding structure in time},
  author={Elman, Jeffrey L},
  journal={Cognitive science},
  volume={14},
  number={2},
  pages={179--211},
  year={1990},
  publisher={Wiley Online Library}
}

@inproceedings{devlin2019bert,
  title={Bert: Pre-training of deep bidirectional transformers for language understanding},
  author={Devlin, Jacob and Chang, Ming-Wei and Lee, Kenton and Toutanova, Kristina},
  booktitle={Proceedings of the 2019 conference of the North American chapter of the association for computational linguistics: human language technologies, volume 1 (long and short papers)},
  pages={4171--4186},
  year={2019}
}

@article{brown2020language,
  title={Language models are few-shot learners},
  author={Brown, Tom and Mann, Benjamin and Ryder, Nick and Subbiah, Melanie and Kaplan, Jared D and Dhariwal, Prafulla and Neelakantan, Arvind and Shyam, Pranav and Sastry, Girish and Askell, Amanda and others},
  journal={Advances in neural information processing systems},
  volume={33},
  pages={1877--1901},
  year={2020}
}

@article{radford2019language,
  title={Language models are unsupervised multitask learners},
  author={Radford, Alec and Wu, Jeffrey and Child, Rewon and Luan, David and Amodei, Dario and Sutskever, Ilya and others},
  journal={OpenAI blog},
  volume={1},
  number={8},
  pages={9},
  year={2019}
}

@article{dosovitskiy2020image,
  title={An image is worth 16x16 words: Transformers for image recognition at scale},
  author={Dosovitskiy, Alexey and Beyer, Lucas and Kolesnikov, Alexander and Weissenborn, Dirk and Zhai, Xiaohua and Unterthiner, Thomas and Dehghani, Mostafa and Minderer, Matthias and Heigold, Georg and Gelly, Sylvain and others},
  journal={arXiv preprint arXiv:2010.11929},
  year={2020}
}

@inproceedings{liu2021swin,
  title={Swin transformer: Hierarchical vision transformer using shifted windows},
  author={Liu, Ze and Lin, Yutong and Cao, Yue and Hu, Han and Wei, Yixuan and Zhang, Zheng and Lin, Stephen and Guo, Baining},
  booktitle={Proceedings of the IEEE/CVF international conference on computer vision},
  pages={10012--10022},
  year={2021}
}

@inproceedings{katharopoulos2020transformers,
  title={Transformers are rnns: Fast autoregressive transformers with linear attention},
  author={Katharopoulos, Angelos and Vyas, Apoorv and Pappas, Nikolaos and Fleuret, Fran{\c{c}}ois},
  booktitle={International conference on machine learning},
  pages={5156--5165},
  year={2020},
  organization={PMLR}
}

@article{choromanski2020rethinking,
  title={Rethinking attention with performers},
  author={Choromanski, Krzysztof and Likhosherstov, Valerii and Dohan, David and Song, Xingyou and Gane, Andreea and Sarlos, Tamas and Hawkins, Peter and Davis, Jared and Mohiuddin, Afroz and Kaiser, Lukasz and others},
  journal={arXiv preprint arXiv:2009.14794},
  year={2020}
}

@article{smith2022simplified,
  title={Simplified state space layers for sequence modeling},
  author={Smith, Jimmy TH and Warrington, Andrew and Linderman, Scott W},
  journal={arXiv preprint arXiv:2208.04933},
  year={2022}
}

@article{biamonte2017quantum,
  title={Quantum machine learning},
  author={Biamonte, Jacob and Wittek, Peter and Pancotti, Nicola and Rebentrost, Patrick and Wiebe, Nathan and Lloyd, Seth},
  journal={Nature},
  volume={549},
  number={7671},
  pages={195--202},
  year={2017},
  publisher={Nature Publishing Group UK London}
}

@article{schuld2015introduction,
  title={An introduction to quantum machine learning},
  author={Schuld, Maria and Sinayskiy, Ilya and Petruccione, Francesco},
  journal={Contemporary Physics},
  volume={56},
  number={2},
  pages={172--185},
  year={2015},
  publisher={Taylor \& Francis}
}

@article{schuld2019quantum,
  title={Quantum machine learning in feature Hilbert spaces},
  author={Schuld, Maria and Killoran, Nathan},
  journal={Physical review letters},
  volume={122},
  number={4},
  pages={040504},
  year={2019},
  publisher={APS}
}

@article{cerezo2021variational,
  title={Variational quantum algorithms},
  author={Cerezo, Marco and Arrasmith, Andrew and Babbush, Ryan and Benjamin, Simon C and Endo, Suguru and Fujii, Keisuke and McClean, Jarrod R and Mitarai, Kosuke and Yuan, Xiao and Cincio, Lukasz and others},
  journal={Nature Reviews Physics},
  volume={3},
  number={9},
  pages={625--644},
  year={2021},
  publisher={Nature Publishing Group UK London}
}

@article{benedetti2019parameterized,
  title={Parameterized quantum circuits as machine learning models},
  author={Benedetti, Marcello and Lloyd, Erika and Sack, Stefan and Fiorentini, Mattia},
  journal={Quantum science and technology},
  volume={4},
  number={4},
  pages={043001},
  year={2019},
  publisher={IOP Publishing}
}

@article{schuld2020circuit,
  title={Circuit-centric quantum classifiers},
  author={Schuld, Maria and Bocharov, Alex and Svore, Krysta M and Wiebe, Nathan},
  journal={Physical Review A},
  volume={101},
  number={3},
  pages={032308},
  year={2020},
  publisher={APS}
}

@article{benedetti2019generative,
  title={A generative modeling approach for benchmarking and training shallow quantum circuits},
  author={Benedetti, Marcello and Garcia-Pintos, Delfina and Perdomo, Oscar and Leyton-Ortega, Vicente and Nam, Yunseong and Perdomo-Ortiz, Alejandro},
  journal={npj Quantum information},
  volume={5},
  number={1},
  pages={45},
  year={2019},
  publisher={Nature Publishing Group UK London}
}

@article{chen2022variational,
  title={Variational quantum reinforcement learning via evolutionary optimization},
  author={Chen, Samuel Yen-Chi and Huang, Chih-Min and Hsing, Chia-Wei and Goan, Hsi-Sheng and Kao, Ying-Jer},
  journal={Machine Learning: Science and Technology},
  volume={3},
  number={1},
  pages={015025},
  year={2022},
  publisher={IOP Publishing}
}

@article{beer2020training,
  title={Training deep quantum neural networks},
  author={Beer, Kerstin and Bondarenko, Dmytro and Farrelly, Terry and Osborne, Tobias J and Salzmann, Robert and Scheiermann, Daniel and Wolf, Ramona},
  journal={Nature communications},
  volume={11},
  number={1},
  pages={808},
  year={2020},
  publisher={Nature Publishing Group UK London}
}

@article{li2023quantum,
  title={Quantum recurrent neural networks for sequential learning},
  author={Li, Yanan and Wang, Zhimin and Han, Rongbing and Shi, Shangshang and Li, Jiaxin and Shang, Ruimin and Zheng, Haiyong and Zhong, Guoqiang and Gu, Yongjian},
  journal={Neural Networks},
  volume={166},
  pages={148--161},
  year={2023},
  publisher={Elsevier}
}

@article{child2019generating,
  title={Generating long sequences with sparse transformers},
  author={Child, Rewon and Gray, Scott and Radford, Alec and Sutskever, Ilya},
  journal={arXiv preprint arXiv:1904.10509},
  year={2019}
}

@article{tay2022efficient,
  title={Efficient transformers: A survey},
  author={Tay, Yi and Dehghani, Mostafa and Bahri, Dara and Metzler, Donald},
  journal={ACM Computing Surveys},
  volume={55},
  number={6},
  pages={1--28},
  year={2022},
  publisher={ACM New York, NY}
}

@article{dao2022flashattention,
  title={Flashattention: Fast and memory-efficient exact attention with io-awareness},
  author={Dao, Tri and Fu, Dan and Ermon, Stefano and Rudra, Atri and R{\'e}, Christopher},
  journal={Advances in neural information processing systems},
  volume={35},
  pages={16344--16359},
  year={2022}
}

@article{gu2021combining,
  title={Combining recurrent, convolutional, and continuous-time models with linear state space layers},
  author={Gu, Albert and Johnson, Isys and Goel, Karan and Saab, Khaled and Dao, Tri and Rudra, Atri and R{\'e}, Christopher},
  journal={Advances in neural information processing systems},
  volume={34},
  pages={572--585},
  year={2021}
}

@book{nielsen2010quantum,
  title={Quantum computation and quantum information},
  author={Nielsen, Michael A and Chuang, Isaac L},
  year={2010},
  publisher={Cambridge university press}
}

@article{zaheer2020big,
  title={Big bird: Transformers for longer sequences},
  author={Zaheer, Manzil and Guruganesh, Guru and Dubey, Kumar Avinava and Ainslie, Joshua and Alberti, Chris and Ontanon, Santiago and Pham, Philip and Ravula, Anirudh and Wang, Qifan and Yang, Li and others},
  journal={Advances in neural information processing systems},
  volume={33},
  pages={17283--17297},
  year={2020}
}

@article{lloyd2013quantum,
  title={Quantum algorithms for supervised and unsupervised machine learning},
  author={Lloyd, Seth and Mohseni, Masoud and Rebentrost, Patrick},
  journal={arXiv preprint arXiv:1307.0411},
  year={2013}
}

@article{lloyd2014quantum,
  title={Quantum principal component analysis},
  author={Lloyd, Seth and Mohseni, Masoud and Rebentrost, Patrick},
  journal={Nature Physics},
  volume={10},
  number={9},
  pages={631--633},
  year={2014},
  publisher={Nature Publishing Group}
}

@article{schuld2016prediction,
  title={Prediction by linear regression on a quantum computer},
  author={Schuld, Maria and Sinayskiy, Ilya and Petruccione, Francesco},
  journal={Physical Review A},
  volume={94},
  number={2},
  pages={022342},
  year={2016},
  publisher={APS}
}

@article{kerenidis2020quantum,
  title={Quantum gradient descent for linear systems and least squares},
  author={Kerenidis, Iordanis and Prakash, Anupam},
  journal={Physical Review A},
  volume={101},
  number={2},
  pages={022316},
  year={2020},
  publisher={APS}
}

@article{rebentrost2014quantum,
  title={Quantum support vector machine for big data classification},
  author={Rebentrost, Patrick and Mohseni, Masoud and Lloyd, Seth},
  journal={Physical review letters},
  volume={113},
  number={13},
  pages={130503},
  year={2014},
  publisher={APS}
}

@article{huang2021experimental,
  title={Experimental quantum generative adversarial networks for image generation},
  author={Huang, He-Liang and Du, Yuxuan and Gong, Ming and Zhao, Youwei and Wu, Yulin and Wang, Chaoyue and Li, Shaowei and Liang, Futian and Lin, Jin and Xu, Yu and others},
  journal={Physical Review Applied},
  volume={16},
  number={2},
  pages={024051},
  year={2021},
  publisher={APS}
}

@article{cong2019quantum,
  title={Quantum convolutional neural networks},
  author={Cong, Iris and Choi, Soonwon and Lukin, Mikhail D},
  journal={Nature Physics},
  volume={15},
  number={12},
  pages={1273--1278},
  year={2019},
  publisher={Nature Publishing Group UK London}
}

@article{romero2017quantum,
  title={Quantum autoencoders for efficient compression of quantum data},
  author={Romero, Jonathan and Olson, Jonathan P and Aspuru-Guzik, Alan},
  journal={Quantum Science and Technology},
  volume={2},
  number={4},
  pages={045001},
  year={2017},
  publisher={IOP Publishing}
}

@article{panella2011neural,
  title={Neural networks with quantum architecture and quantum learning},
  author={Panella, Massimo and Martinelli, Giuseppe},
  journal={International Journal of Circuit Theory and Applications},
  volume={39},
  number={1},
  pages={61--77},
  year={2011},
  publisher={Wiley Online Library}
}

@article{mitarai2018quantum,
  title={Quantum circuit learning},
  author={Mitarai, Kosuke and Negoro, Makoto and Kitagawa, Masahiro and Fujii, Keisuke},
  journal={Physical Review A},
  volume={98},
  number={3},
  pages={032309},
  year={2018},
  publisher={APS}
}

@article{nguyen2023quantum,
  title={Quantum vision clustering},
  author={Nguyen, Xuan Bac and Churchill, Hugh and Luu, Khoa and Khan, Samee U},
  journal={arXiv preprint arXiv:2309.09907},
  year={2023}
}

@article{nguyen2025diffusion,
  title={Diffusion-inspired quantum noise mitigation in parameterized quantum circuits},
  author={Nguyen, Hoang-Quan and Nguyen, Xuan Bac and Chen, Samuel Yen-Chi and Churchill, Hugh and Borys, Nicholas and Khan, Samee U and Luu, Khoa},
  journal={Quantum Machine Intelligence},
  volume={7},
  number={1},
  pages={55},
  year={2025},
  publisher={Springer}
}

@inproceedings{nguyen2024hierarchical,
  title={Hierarchical quantum control gates for functional mri understanding},
  author={Nguyen, Xuan-Bac and Nguyen, Hoang-Quan and Churchill, Hugh and Khan, Samee U and Luu, Khoa},
  booktitle={2024 IEEE Workshop on Signal Processing Systems (SiPS)},
  pages={159--164},
  year={2024},
  organization={IEEE}
}

@article{nguyen2024quantum,
  title={Quantum visual feature encoding revisited},
  author={Nguyen, Xuan-Bac and Nguyen, Hoang-Quan and Churchill, Hugh and Khan, Samee U and Luu, Khoa},
  journal={Quantum Machine Intelligence},
  volume={6},
  number={2},
  pages={61},
  year={2024},
  publisher={Springer}
}

@inproceedings{nguyen2024qclusformer,
  title={Qclusformer: A quantum transformer-based framework for unsupervised visual clustering},
  author={Nguyen, Xuan-Bac and Nguyen, Hoang-Quan and Chen, Samuel Yen-Chi and Khan, Samee U and Churchill, Hugh and Luu, Khoa},
  booktitle={2024 IEEE International Conference on Quantum Computing and Engineering (QCE)},
  volume={2},
  pages={347--352},
  year={2024},
  organization={IEEE}
}

@article{nguyen2024quantum_brain,
  title={Quantum-brain: Quantum-inspired neural network approach to vision-brain understanding},
  author={Nguyen, Hoang-Quan and Nguyen, Xuan-Bac and Churchill, Hugh and Choudhary, Arabinda Kumar and Sinha, Pawan and Khan, Samee U and Luu, Khoa},
  journal={arXiv preprint arXiv:2411.13378},
  year={2024}
}

@inproceedings{nguyen2025qmoe,
  title={Qmoe: A quantum mixture of experts framework for scalable quantum neural networks},
  author={Nguyen, Hoang-Quan and Nguyen, Xuan-Bac and Pandey, Sankalp and Khan, Samee U and Safro, Ilya and Luu, Khoa},
  booktitle={2025 IEEE International Conference on Quantum Computing and Engineering (QCE)},
  volume={2},
  pages={223--228},
  year={2025},
  organization={IEEE}
}

@inproceedings{holliday2025quadro,
  title={QUADRO: A Hybrid Quantum Optimization Framework for Drone Delivery},
  author={Holliday, James B and Blount, Darren and Nguyen, Hoang Quan and Khan, Samee U and Luu, Khoa},
  booktitle={2025 IEEE International Conference on Quantum Computing and Engineering (QCE)},
  volume={1},
  pages={2090--2100},
  year={2025},
  organization={IEEE}
}

@article{farhi2014quantum,
  title={A quantum approximate optimization algorithm},
  author={Farhi, Edward and Goldstone, Jeffrey and Gutmann, Sam},
  journal={arXiv preprint arXiv:1411.4028},
  year={2014}
}

@article{zhou2020quantum,
  title={Quantum approximate optimization algorithm: Performance, mechanism, and implementation on near-term devices},
  author={Zhou, Leo and Wang, Sheng-Tao and Choi, Soonwon and Pichler, Hannes and Lukin, Mikhail D},
  journal={Physical Review X},
  volume={10},
  number={2},
  pages={021067},
  year={2020},
  publisher={APS}
}

@article{pandey2026openqlaw,
  title={OpenQlaw: An Agentic AI Assistant for Analysis of 2D Quantum Materials},
  author={Pandey, Sankalp and Nguyen, Xuan-Bac and Nguyen, Hoang-Quan and Faltermeier, Tim and Borys, Nicholas and Churchill, Hugh and Luu, Khoa},
  journal={arXiv preprint arXiv:2603.17043},
  year={2026}
}

@inproceedings{nguyen2026qupaint,
  title={QuPAINT: Physics-Aware Instruction Tuning Approach to Quantum Material Discovery},
  author={Nguyen, Xuan-Bac and Nguyen, Hoang-Quan and Pandey, Sankalp and Faltermeier, Tim and Borys, Nicholas and Churchill, Hugh and Luu, Khoa},
  booktitle={Proceedings of the IEEE/CVF Conference on Computer Vision and Pattern Recognition},
  year={2026}
}

@article{nguyen2025phi_adapt,
  title={Phi-Adapt: A Physics-Informed Adaptation Learning Approach to 2D Quantum Material Discovery},
  author={Nguyen, Hoang-Quan and Nguyen, Xuan Bac and Pandey, Sankalp and Faltermeier, Tim and Borys, Nicholas and Churchill, Hugh and Luu, Khoa},
  journal={arXiv preprint arXiv:2507.05184},
  year={2025}
}
}

\end{document}